%% file: main.tex
\documentclass[runningheads]{llncs}

\input{macro.tex}

\title{DeepOPG: Improving Orthopantomogram Finding Summarization with Weak Supervision}
\titlerunning{DeepOPG: Improving OPG Finding Summarization with Weak Supervision}

\author{
    Tzu-Ming Harry Hsu\inst{1} \and
    Yin-Chih Chelsea Wang\inst{2}}

\institute{
    MIT CSAIL, USA\\\email{stmharry@mit.edu} \and
    Chung Shan Medical University, Taiwan\\\email{chelsea850416@gmail.com}}

\begin{document}
\maketitle

\input{0-abstract}
\input{1-introduction}
\input{2-methods}
\input{3-experiments}
\input{4-conclusion}

\bibliographystyle{unsrtnat}
\bibliography{main}

\newpage
\section*{Supplementary Material}
\input{appendix.tex}

\end{document}

%% file: macro.tex
\usepackage{amsmath}
\usepackage{amssymb}
\usepackage{array}
\usepackage{bbm}
\usepackage{booktabs}
\usepackage[inline]{enumitem}
\usepackage{graphicx}
\usepackage{hhline}
\usepackage[pagebackref=true,breaklinks=true,letterpaper=true,colorlinks,bookmarks=false]{hyperref}
\usepackage{multirow}
\usepackage[square,numbers,sectionbib]{natbib}
\usepackage{subcaption}
\usepackage{tablefootnote}
\usepackage{url}
\usepackage{xcolor}
\usepackage{xspace}

\renewcommand\subsubsection[1]{\noindent\textbf{#1}\enspace}

\newcommand{\opg}{OPG\xspace}
\newcommand{\opgfull}{orthopantomogram\xspace}
\newcommand{\this}{DeepOPG\xspace}

\renewcommand{\b}{{\mathbf{b}}}
\newcommand{\e}{{\mathbf{e}}}
\newcommand{\E}{{\mathbf{E}}}
\renewcommand{\L}{{\mathcal{L}}}
\newcommand{\p}{{\mathbf{p}}}
\renewcommand{\P}{{\mathbf{P}}}
\newcommand{\R}{{\mathbb{R}}}
\newcommand{\M}{{\mathbf{M}}}
\renewcommand{\det}{{\mathrm{det}}}
\newcommand{\seg}{{\mathrm{seg}}}
\newcommand{\dcr}{{\mathrm{DCR}}}

\newcommand{\ind}{\mathbbm{1}}
\renewcommand{\exp}{\mathbb{E}}
\newcommand{\tp}{\mathrm{TP}}
\newcommand{\fp}{\mathrm{FP}}
\newcommand{\fn}{\mathrm{FN}}
\newcommand{\tn}{\mathrm{TN}}
\newcommand{\iou}{\mathrm{IoU}}
\newcommand{\gt}{\mathrm{gt}}

\newcommand{\para}[1]{\left(#1\right)}
\newcommand{\param}[1]{\left[#1\right]}
\newcommand{\paral}[1]{\left\{#1\right\}}
\newcommand{\abs}[1]{\left|#1\right|}

\makeatletter
\DeclareRobustCommand\onedot{\futurelet\@let@token\@onedot}
\def\@onedot{\ifx\@let@token.\else.\null\fi\xspace}
\def\ie{\emph{i.e}\onedot}

\newcolumntype{L}[1]{>{\raggedright\let\newline\\\arraybackslash}m{#1}}
\newcolumntype{C}[1]{>{\centering\let\newline\\\arraybackslash}m{#1}}
\newcolumntype{R}[1]{>{\raggedleft\let\newline\\\arraybackslash}m{#1}}
\newcommand{\makecell}[2][c]{\begin{tabular}[c]{@{}#1@{}}#2\end{tabular}}

\makeatletter
\newcommand\footnoteref[1]{\protected@xdef\@thefnmark{\ref{#1}}\@footnotemark}
\makeatother

%% file: 0-abstract.tex
\begin{abstract}
Clinical finding summaries from an orthopantomogram, or a dental panoramic radiograph, have significant potential to improve patient communication and speed up clinical judgments. 
While orthopantomogram is a first-line tool for dental examinations, no existing work has explored the summarization of findings from it. 
A finding summary has to find teeth in the imaging study and label the teeth with several types of past treatments.
To tackle the problem, we develop \this that breaks the summarization process into functional segmentation and tooth localization, the latter of which is further refined by a novel dental coherence module.
We also leverage weak supervision labels to improve detection results in a reinforcement learning scenario.
Experiments show high efficacy of \this on finding summarization, achieving an overall AUC of 88.2\% in detecting six types of findings.
The proposed dental coherence and weak supervision are shown to improve DeepOPG by adding 5.9\% and 0.4\% to AP@IoU=0.5.


\keywords{
    Orthopantomogram \and 
    Dental Panoramic Radiograph \and
    Reinforcement Learning \and
    Weak Supervision
}
\end{abstract}

%% file: 1-introduction.tex
\section{Introduction}
An orthopantomogram (OPG), or a dental panoramic radiograph, is a half-circle X-ray scanning of the oral region that compresses the complicated 3D structures to a 2D representation as shown in Figure~\ref{fig:overview}.
\opg has many advantages including short acquisition time and convenience of examination.
Moreover, its capability to deliver rich information about the oral and maxillofacial regions makes it a first-line dental screening tool~\cite{perschbacher2012interpretation}.
With that said, it is this structural complexity that unavoidably limits the interpretation of \opg to only dental experts~\cite{henzler2018single}.
Even for these dental experts, interpretation of findings can suffer from insufficient inter-rater agreement~\cite{kweon2018panoramic} and low time efficiency in clinical practices~\cite{plessas2019impact,rozylo2018artificial}.
As such, an automatic system to provide finding summaries on the fly can be beneficial in terms of both patient communication and clinical assistance.
The systematically collected summaries can further provide an invaluable source for subsequent dental research and statistical analysis, which the current clinical workflow cannot offer.

There have been attempts to provide information about teeth in radiographs with convolutional neural networks (CNNs).
In \cite{ronneberger2015dental}, they offer pixel-wise segmentation maps that label seven different parts of teeth.
\cite{miki2017classification} classifies teeth images into eight categories but requires that the bounding boxes be manually annotated first.
\cite{koch2019accurate} identifies silhouettes for natural teeth in \opg with semantic segmentation but treats all teeth as a single connected region.
\cite{silva2018automatic,jader2018deep} use a novel \opg dataset and object detection to treat teeth as individual instances for object detection, yet they both do not number the teeth.
\cite{tuzoff2019tooth} addresses both detection and numbering, but fails to include dental implants.
\cite{kim2020automatic} provides detection of teeth, implant, and crowns but does not associate them with findings.
Moreover, the vast majority of past research relies on annotations on dense attribute maps which is resource-intensive, and the use of weaker (and faster to collect) supervision has not been explored.

In this work, we aim to provide a summary of findings in an \opg image, including all teeth found in the image, their FDI notations, and all the clinical findings on each.
We propose \this, which breaks the finding summarization process into two sub-tasks: functional segmentation and tooth localization, the latter of which is further refined at inference-time by maximizing the novel \emph{Dental Coherence Reward (DCR)}. 
DCR can also be used by reinforcement learning (RL) for training-time optimization, leveraging the \emph{missing teeth annotation} that are quick for dentists to label as weak supervision.
We curate a set of annotations on \opg including semantic segmentation, instance segmentation, and finding summaries for 298 studies on a dataset in the public domain.
Our experiments show that \this achieves an overall AUC of 88.2\% on finding detection, which is 1.6\% higher than without weak supervision.
The tooth/implant localization yields an average precision at zero IoU of 98.6\%, which is 5.6\% higher than without injecting dental domain knowledge and 0.9\% higher than without feeding in segmentation maps.
The numbers demonstrate the effectiveness of each component of \this.
To our knowledge, this is the first work to explore the summarization of findings in \opg images and to use weak supervision to improve finding summarization. 

%% file: 2-methods.tex
\section{Methods}

\begin{figure}[!t]
    \centering
    \begin{subfigure}{0pt}
        \phantomsubcaption\label{fig:overview:a}
    \end{subfigure}
    \begin{subfigure}{0pt}
        \phantomsubcaption\label{fig:overview:b}
    \end{subfigure}
    \begin{subfigure}{0pt}
        \phantomsubcaption\label{fig:overview:c}
    \end{subfigure}
    \includegraphics[width=1.0\textwidth]{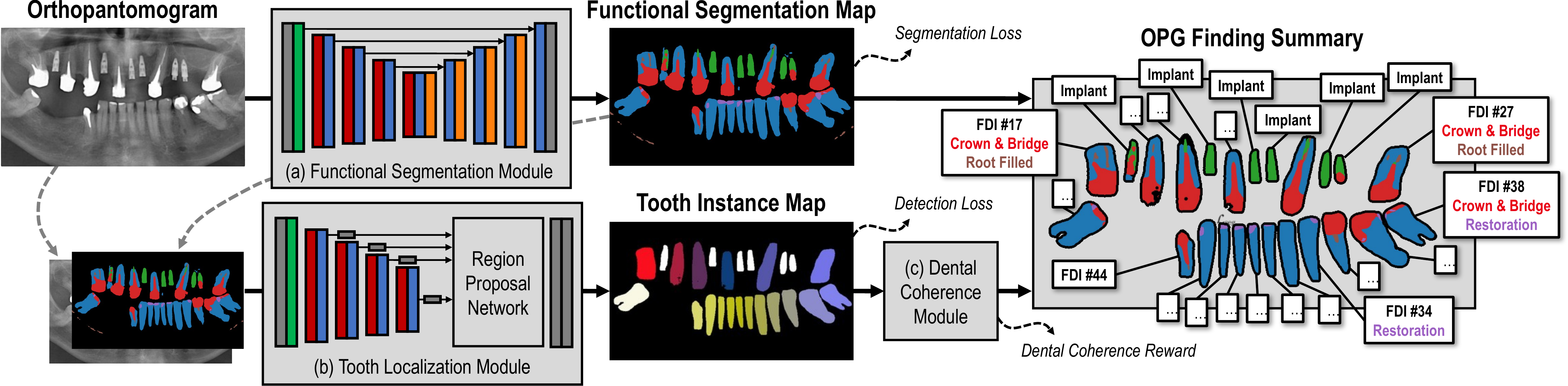}
    \caption{
        \textbf{System Overview of \this.}
        There are three modules in \this working together to obtain \opgfull finding summary on each tooth.
        (a) Functional Segmentation Module performs per-pixel classification,
        (b) Tooth localization Module localizes each tooth, and (c) Dental Coherence Module ensures that output is clinically reasonable. 
        Ultimately, we combine the segmentation maps and teeth identity maps to produce findings with explainable predictive values.}
    \label{fig:overview}
\end{figure}

Our ultimate goal for the \this system is to generate a finding summary entailing six different types of findings on each tooth in a \opg.
The resulting findings are formulated as binary attribute labels on the teeth found in the \opg.
We decompose the problem into two main tasks: localizing the objects of interest (in this case, teeth and implants) and determining the visual features that result in the findings.
Illustrated in Figure~\ref{fig:overview}, there are three modules in \this, and they operate at original resolutions of the images. 
This is essential since some findings (e.g., fillings found in the root canal) are visually tiny, and any down-sampling would result in a loss of information.
We combine the results from both tasks of localization and function determination to output predictive values for each of the finding types on individual teeth.

\subsection{Model Architecture}
First of all, the \emph{functional segmentation} module as shown in Figure~\ref{fig:overview:a} consumes a radiographic image as the input and generates a map that shows the dental functionality of each pixel.
The functional segmentation map and the original image is then concatenated to go into the \emph{tooth localization} module in Figure~\ref{fig:overview:b} that picks out the individual dental object of interest including teeth and implants.
The resulting detection outcomes are further refined by the \emph{dental coherence} module where clinical heuristics are applied to ensure coherence with dental knowledge. 

\subsubsection{Functional Segmentation via Semantic Segmentation}
Given the input gray-scale image $I \in \R^{H \times W \times 1}$, where $H$ and $W$ are the height and width of the image, we employ a network with an U-Net-like~\cite{ronneberger2015dental} structure to predict $S \in \R^{H \times W \times C_\seg}$, a per-class probability for each pixel that determines its functional class $c \in \paral{1, 2, \ldots, C_\seg}$, where $C_\seg$ is the number of classes.
In our experiments, $C_\seg=7$ and includes the following classes for finding summarization:
\begin{enumerate*}[label=(\arabic*)]
    \item background,
    \item normal (non-impacted) teeth,
    \item impaction,
    \item crown \& bridge,
    \item restoration,
    \item root filling material, and
    \item implant.
\end{enumerate*}
Note that as these classes are mutually exclusive, a ground truth segmentation map $S^\gt \in \R^{H \times W \times C_\seg}$ is one-hot encoded and the activation function of network output is thus softmax. 
Specifically, we choose ResNet-50~\cite{he2016deep} to be the encoder and ResNet-18 with transposed convolutions to be the decoder.

\subsubsection{Tooth localization via Object Detection}
The tooth localization module takes the concatenated image $\param{I, S} \in \R^{H \times W \times \para{1 + C_\seg}}$ and produces $N$ \emph{detections}, each including a class probability vector $\p_n \in \R^{C_\det}$, a region of interest (ROI) $\b_n \in \R^4$, and the class-wise masks $M_n \in \R^{H \times W \times C_\det}$, where $C_\det$ is the number of classes in detection.
Concretely, we adopt Mask-RCNN~\cite{he2017mask} that proposes a pool of candidate ROIs with a region proposal network (RPN) before using a small sub-network to derive the aforementioned detection properties.


As we are interested in not only natural teeth but also dental implants, in total there are $C_\det = 34$ classes representing the background, 32 different teeth in permanent dentition, and the implant.
Hereafter the natural teeth are annotated using the FDI World Dental Federation notation as shown in Figure~\ref{fig:quantitative:a}.

\subsubsection{Inference-Time Dental Coherence Decoding}
One major downside of directly using off-the-shelf detection algorithms is that they mostly consider the detection efficacy of individual objects rather than the conglomerate of several objects.
As a result, in pilot experiments, we often observe the detection module to output several objects of interest with the same FDI tooth number, which is highly unlikely in practice.
Equally frequently, there are cases where an image patch can be detected as multiple different classes at the same time, with largely overlapping masks.
Even with existing techniques such as non-maximum suppression (NMS) that filters out overlapping objects with lower confidence scores, we are only able to partially resolve the latter problem.

To this end, we propose to look at this problem from an optimization perspective and \emph{decode} an assignment $\E$ of teeth number to the detected objects by maximizing the Dental Coherence Reward (DCR) defined as 
\begin{equation}
    r_\dcr\para{\P, \E, \M} \equiv \sum_{n,c} p_{nc} \cdot e_{nc} - \sum_{n,c,m,d} q_{ncmd} \cdot e_{nc} \cdot e_{md},
    \label{eqn:dcr}
\end{equation}
subject to 
$
    \sum_n e_{nc} \le 1 \;
    \forall c \in \paral{1, 2, \ldots, C}, 
$
and
$
    e_{nc} \in \paral{0, 1} \;
    \forall c \in \paral{1, 2, \ldots, C}
    \forall n \in \paral{1, 2, \ldots, N},
$
where $p_{nc} = \para{\p_n}_c$ is the probability of object $n$ belonging to class $c$, $q_{ncmd} = \frac{\para{M_n}_c \cap \para{M_m}_d}{ \para{M_n}_c \cup \para{M_m}_d}$ is the intersection-over-union between masks $\para{M_n}_c$ (the class-$c$ mask of object $n$) and $\para{M_m}_d$, and $e_{nc}$ is an indicator whether we \emph{assign} tooth $c$ to object $n$.
Note that an object $n$ can be \emph{suppressed} (\ie, discarded) if $\sum_c e_{nc} = 0$. 
This formulation happens to be the \emph{Generalized Quadratic Assignment Problem} (GQAP)~\cite{lee2004generalized} which is extensively studied in optimization theory and has solvers widely available.
Implants are not modified in this module, and hence $C=C_\det-1$ with implants excluded for optimization.

The idea to maximize DCR closely resembles how our dental experts parse an \opg, where they explain they would
\begin{enumerate*}[label=(\arabic*)]
    \item
        identify all minimally overlapping objects and mentally assign each a number, followed by
    \item 
        ensuring that across a single image, no teeth share the same FDI number (obviously, multiple dental implants can still present simultaneously).
\end{enumerate*}
While it is certainly possible in the clinics to observe the extremely rare cases where two natural teeth overlap on the \opg, oftentimes highly overlapping masks simply indicate that a tooth is independently recognized by two RPN proposals.

\subsubsection{Explainable \opg Finding Summary}
We assemble the information from the semantic segmentation and the detection outputs to derive the finding summary.
For each of the teeth or implants, we use its mask $M$ to select the corresponding regions in the segmentation map $S$ and calculate the percentage of pixel counts for each functional class $c$ in that area as
$
    f_c = \sum_{i \in M} \frac{\ind\param{S_i = c}}{\abs{M}},
$
where $\ind\param{\cdot}$ is the indicator function.
The percentage area $f_c$ is then used as the predictive value for finding type $c$ on that tooth.
Doing so not only allows us to provide an explainable finding output that dentists can easily reason, but we also can adjust the threshold on $f_c$ based on our sensitivity/specificity requirements.

\subsection{Improved Tooth Localization with Weakly Supervised Reinforcement Learning}
The annotation for tooth localization usually requires that the dental experts carefully outline the silhouettes of each tooth and provide an FDI number for it. 
This type of annotation is labor-intensive and is usually not available at most data registries. 
What is more likely to be available is a description of whether a tooth is missing or not in a text report (\ie, $\sum_n e_{nc} = 1$ if the tooth $c$ is present and 0 otherwise). 
We hereby are interested to find if weak supervision in the form of tooth missingness is helpful to train the tooth localization module in a reinforcement learning (RL) scenario.

We utilize the REINFORCE~\cite{williams1992simple} algorithm where as long as a probability and a reward are defined for output, the network can learn to maximize the reward function.
At training time, instead of decoding the GQAP problem, we sample an one-hot vector $\hat\e_n = \param{\hat{e}_{n1}, \hat{e}_{n2}, \ldots, \hat{e}_{nC}} \sim \p_n$ from the class distribution $\p_n$ for each object $n$ independently.
As the random samples might violate the constraint that each FDI number cannot be taken by multiple teeth (\ie, $\sum_n e_{nc} > 1$), we penalize this situation by setting the reward for extra teeth to be negative
\begin{equation}
    \footnotesize
    \hat{p}_{nc} =
    \begin{cases}
        + p_{nc} & \text{if tooth $c$ is present and $p_{nc}$ is the largest probability for it} \\
        - p_{nc} & \text{otherwise (for extra teeth),}
    \end{cases}
\end{equation}
and calculate $r_\dcr\para{\hat\P,\hat\E,\M}$ on the samples.
The loss as given by REINFORCE is thus
\begin{equation}
    \L_\dcr = -\exp_{\hat\E \sim p_\theta\para{\hat\E}}\param{
        r_\dcr\para{\hat\P,\hat\E,\M}
        \sum_{n,c} \hat{e}_{nc} \log \hat{p}_{nc}
    },
\end{equation}
where $p_\theta\para{\cdot}$ is the distribution characterized by the network.
We can approximate the above gradient with Monte-Carlo samples and average gradients across training examples in the batch.
Different from the aforementioned inference-time decoding, we can explicitly optimize the network for DCR with reinforcement learning here.

To learn \this, we employ a multi-stage learning procedure since the RPN in Mask-RCNN is non-differentiable. 
First, we train the functional segmentation module, optimizing the segmentation cross-entropy loss $\L_\seg$.
Following this, the tooth localization module learns using the inference-time predicted segmentation maps and minimizes a loss $\L_\det$ as detailed in~\cite{he2017mask}.
Finally, we fine-tune the tooth localization module with DCR weak supervision, minimizing the joint loss $\L=\L_\det + \L_\dcr$, freezing all network layers except for the last.
For implementation details, please refer to the supplementary material.

%% file: 3-experiments.tex
\section{Experiments}
In this section, we provide validation of individual modules as well as \this as a whole. 
First of all, we present a dataset with novel annotations on segmentation, detection, and finding summary.
We then offer an overview of the finding summarization efficacy for each of the finding types.
Following this, we provide an ablation study on the tooth localization module including our proposed DCR decoding and reinforcement learning.
Finally, we compare our \this with existing works under comparable settings.
For brevity, the performance of the functional segmentation module is provided in the supplementary material.

\subsection{Dataset}
In this work, we use the UFBA-UESC Dental Images Deep dataset~\cite{silva2018automatic} where there are 1,500 \opg images in total, 267 out of which are annotated for tooth localization (implant annotations are not provided in the original data).
The \opg images can be split into four major categories:
\begin{enumerate*}[label=(\arabic*)]
    \item studies with all permanent dentition present and no implants,
    \item studies with missing teeth and no implants,
    \item studies with implants, and
    \item studies with mixed dentition.
\end{enumerate*}
We exclude all studies with mixed dentition and supernumerary teeth as they are outside the scope of this work.

To enrich the dataset for learning \this, we ask 3 board-certified dentists to provide additional annotations including 
\begin{enumerate*}[label=(\arabic*)]
    \item functional segmentation maps on 68 studies,
    \item tooth/implant localization maps on 39 studies,
    \item tooth/implant missingness summary (weak supervision, in the form of 32 binary labels per study) on 144 studies, and
    \item finding summary (in the form of 32 $\times$ 6 binary labels per study) on 47 studies.
\end{enumerate*}

To avoid overfitting the data, no study is annotated for two or more annotation types.
It is important to note that segmentation/localization maps take, on average, 30 minutes to annotate per study, while the teeth missingness information only takes 30 seconds each.
In each stage of \this learning, data is split into 70/30 training/test randomly, and the finding summary is exclusively used as test data.

\subsection{Overall Evaluation of \this for Findings Summarization}

\begin{table}[t]
\centering
\caption{ \textbf{AUC Comparisons.} We compare the AUCROC for six \opg finding types for two settings of \this.  See Section~\ref{sec:localization} for method descriptions. }
\label{tbl:overall}
\footnotesize
\begin{tabular}{ L{2.3cm}C{1.3cm}C{1.4cm}C{1.4cm}C{1.4cm}C{1.1cm}C{1.4cm}|C{1.0cm} }
\toprule
\multirow{2}{*}[-1.0em]{\textbf{ Method }} & \multicolumn{7}{c}{\bf AUCROC (\%) } \\ \cmidrule{2-8}
  & \makecell[c]{ Missing\\ Teeth } & \makecell[c]{ Impacted\\ Teeth } & \makecell[c]{ w/Crown\\ \& Bridge } & \makecell[c]{ w/Resto-\\ ration } & \makecell[c]{ Root\\ Filled } & \makecell[c]{ Implants } & \makecell[c]{ Macro\\ Avg. } \\ \midrule
DeepOPG (full) & \bf 90.6 & \bf 96.9 & 86.5 & \bf 89.3 & \bf 88.2 & 77.6 & \bf 88.2 \\ \hline
w/o RL & 87.6 & 96.5 & \bf 88.2 & 86.4 & 82.9 & \bf 78.1 & 86.6 \\ \hhline{*{8}{=}}
\makecell[l]{ Area Threshold\\ @ max F1 } & 24.2\% & 34.5\% & 25.9\% & 2.70\% & 0.33\% & $-$ & $-$ \\ \bottomrule
\end{tabular}
\end{table}

As mentioned before, \this combines the functional segmentation map and the tooth localization results by calculating the percentage area of each functional class for each tooth.
Using the percentage area as the predictive value for the binary finding labels, we are able to evaluate the overall performance of \this by calculating the receiver operating characteristic (ROC) curve where we plot the true positive rate $\mathrm{TPR} = \frac{\tp}{\tp + \fn}$ against the true negative rate $\mathrm{TNR} = \frac{\tn}{\tn + \fp}$.
Note that for a finding prediction to be $\tp$, it not only has to have enough pixels of that finding in the tooth, but the tooth number itself has to be correctly detected.

The ROC curves for the six types of findings are shown in detail in the supplementary material.
We can calculate the area under curve (AUC) for each of the findings as summarized in Table~\ref{tbl:overall}.
In the table, we also compare a setting where the RL with DCR is disabled. 
It is clear that the weak supervisions with RL can improve the finding summarization.
Of the six findings, impacted teeth with an AUC of $96.9\%$ is the easiest task, possibly because it is a large object and that is often found at fixed locations such as the wisdom teeth.
We also show, on the last row, the threshold on the percentage area at the operating point with the largest $\mathrm{F1} = \frac{2 \times \tp}{2 \times \tp + \fn + \fp}$. 
It is interesting to see that root-filled teeth only require 0.33\% of the area to be finding-positive while impacted teeth require 34.5\% of the tooth to be labeled impacted.

To highlight the usefulness of weak supervision, the ``w/o RL" model (86.6\% AUC) trains with 273 per-pixel annotations which take 136 expert hours to prepare. 
The ``DeepOPG (full)" model adds 100 weak supervision annotations which only take an additional 0.8 expert hours, but a gain of 1.6\% overall AUC. 
This demonstrates weak supervision is effective in boosting AUC while requiring substantially less expert effort (<1\% extra time) than per-pixel annotations.

\subsection{Tooth Localization with Dental Coherence}
\label{sec:localization}

\begin{table}[t]
\centering
\caption{ \textbf{Comparisons of Detection Metrics.} We show detection metrics for various settings of \this.  We report the metric values and their standard errors. $\text{AP}_x$ denotes $\text{AP@IoU}=x$. See Section~\ref{sec:localization} for method descriptions. }
\label{tbl:detection}
\footnotesize
\begin{tabular}{ L{2.5cm}C{1.8cm}C{1.8cm}C{1.6cm}C{1.6cm}C{2.0cm} }
\toprule
\multirow{2}{*}[-0.3em]{\textbf{ Method}} & \multicolumn{4}{c}{\bf Per-Object} & \bf Per-Image \\ \cmidrule{2-5} \cmidrule(l){6-6}
  & $\text{AP}_{0.0}$ (\%) & $\text{AP}_{0.5}$ (\%) & DA (\%) & FA (\%) & IoU (\%) \\ \midrule
DeepOPG (full) & $\mathbf{98.6_{0.1}}$ & $\mathbf{97.6_{0.3}}$ & $\mathbf{98.7_{0.4}}$ & $\mathbf{97.5_{0.6}}$ & $\mathbf{80.5_{1.5}}$ \\ \hline
w/o RL & $98.4_{0.1}$ & $97.2_{0.4}$ & $\mathbf{98.7_{0.4}}$ & $\mathbf{97.5_{0.6}}$ & $80.1_{1.5}$ \\ \hline
\makecell[l]{ w/o RL and \\ dental coherence } & $93.0_{0.1}$ & $91.3_{0.4}$ & $93.7_{0.9}$ & $87.4_{1.2}$ & $79.7_{1.6}$ \\ \hline
\makecell[l]{ w/o segmentation } & $97.7_{0.1}$ & $96.2_{0.3}$ & $97.9_{0.5}$ & $95.7_{0.8}$ & $80.2_{1.5}$ \\ \bottomrule
\end{tabular}
\end{table}

\begin{figure}[!t]
    \centering
    \begin{subfigure}{0pt}
        \phantomsubcaption\label{fig:quantitative:a}
    \end{subfigure}
    \begin{subfigure}{0pt}
        \phantomsubcaption\label{fig:quantitative:b}
    \end{subfigure}
    \begin{subfigure}{0pt}
        \phantomsubcaption\label{fig:quantitative:c}
    \end{subfigure}
    \begin{subfigure}{0pt}
        \phantomsubcaption\label{fig:quantitative:d}
    \end{subfigure}
    \begin{subfigure}{0pt}
        \phantomsubcaption\label{fig:quantitative:e}
    \end{subfigure}
    \begin{subfigure}{0pt}
        \phantomsubcaption\label{fig:quantitative:f}
    \end{subfigure}
    \includegraphics[width=1.0\textwidth]{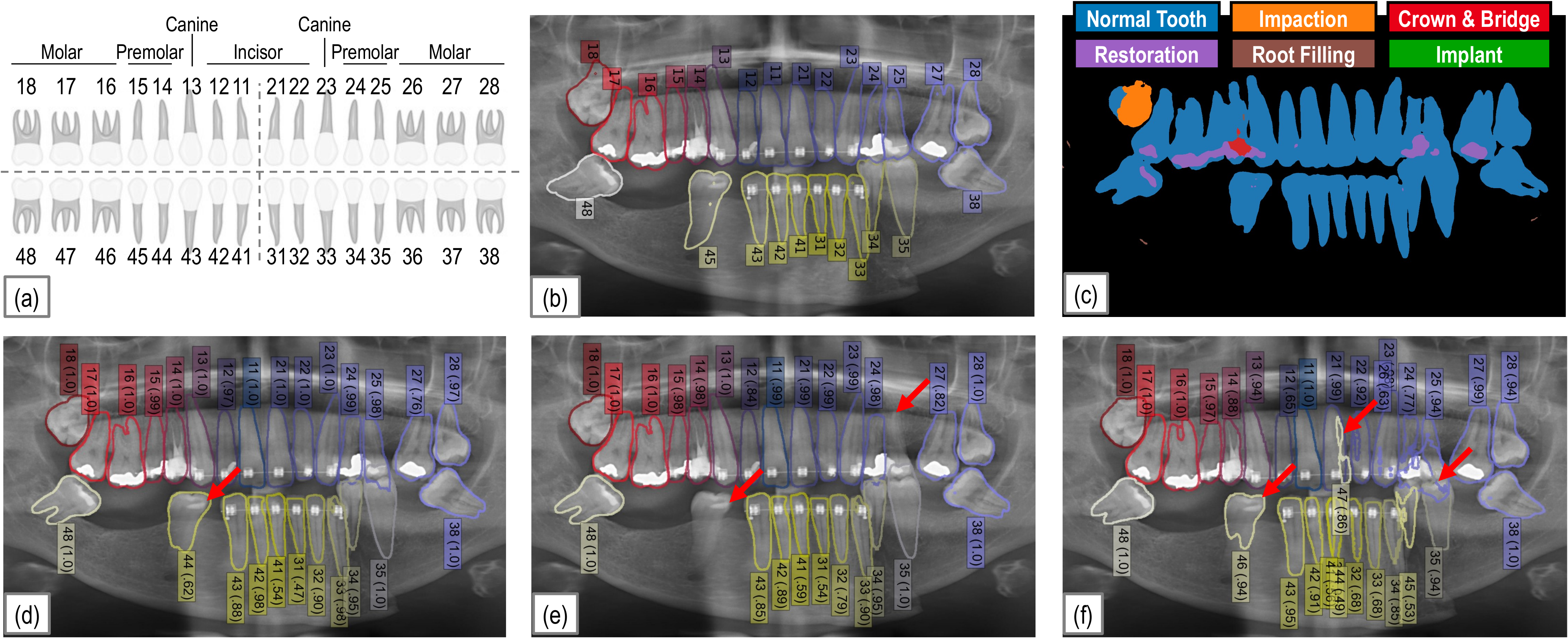}
    \caption{
        \textbf{Illustrations of Tooth localization Results.}
        (a) The FDI tooth numbering system,
        (b) input \opg with ground truth localization,
        (c) functional segmentation map,
        (d) localization for \this (full),
        (e) localization w/o reinforcement learning, and
        (f) localization w/o segmentation input.
    }
    \label{fig:quantitative}
\end{figure}

To verify the efficacy of the proposed modifications to the off-the-shelf object detection networks, we perform several ablation studies to inspect the contribution of these modifications.
In particular, we assess
\begin{enumerate}
    \item \textbf{\this (full)}: We enable all model features, including feeding segmentation maps as the input for the tooth localization, using dental coherence module at inference, and training the model with reinforcement learning.
    \item \textbf{w/o RL}: All model features, except training with RL.
    \item \textbf{w/o RL and dental coherence}: We remove both the dental coherence module and the reinforcement learning components.
    \item \textbf{w/o segmentation}: Segmentation maps are not fed into the tooth localization module in this case.
\end{enumerate}

\subsubsection{Metrics}
The performance of different models are compared using various metrics, including the commonly used average precision (AP) defined in PASCAL VOC~\cite{everingham2010pascal} for detection tasks, the detection accuracy $\mathrm{DA} \equiv \frac{\tp + \fn}{\tp + \fn + \fp}$ and identification accuracy $\mathrm{FA} \equiv \frac{\tp}{\tp + \fn + \fp}$~\cite{cui2019toothnet}.
On a per-image level, we evaluate the intersection over union as $\iou \equiv \frac{\sum_n M_n \cap M_n^\gt}{\sum_n M_n \cup M_n^\gt}$.

In Table~\ref{tbl:detection}, we observe consistent gains in performance across all metrics when we incorporate different proposed features. 
Most notably, the dental coherence module constitutes most of the gains, providing $+5.6\%$ in AP@IoU=0.0 and $+5.0\%$ in DA. 
Using segmentation maps also provides $+0.9\%$ gain in AP@IoU=0.0 since segmentation maps carry more global information by nature.
The weak supervision, while seemingly provides less compelling improvements, is, in fact, remarkable as annotating the teeth missingness summary is faster than annotating the localization maps by orders of magnitudes. 

Figure~\ref{fig:quantitative} showcases localization results on a test image with three different configurations.
This study contains a maloccluded tooth, on which all three configurations predict incorrectly.
It is also worth noting that by removing the segmentation input, the localization depends totally on the input \opg and can be over-sensitive as indicated by red arrows in Figure~\ref{fig:quantitative:f}.

\begin{table}[t]
\centering
\caption{ \textbf{Comparison to Prior Works.} We compare our full model DeepOPG under similar conditions to prior works on various tasks. Note the evaluations are done on different dataset in each work. We report the metric values and their standard errors. $\text{AP}_x$ denotes $\text{AP@IoU}=x$. }
\label{tbl:comparison}
\footnotesize
\begin{tabular}{ L{2.5cm}C{2.0cm}C{1.5cm}C{1.5cm} }
\toprule
\multirow{2}{*}[-0.3em]{\textbf{ Method}} & \multicolumn{3}{c}{\bf Tooth Segmentation} \\ \cmidrule{2-4}
  & Precision (\%) & Recall (\%) & F1 (\%) \\ \midrule
\citet{wirtz2018automatic} & $79.0$ & $82.7$ & $80.3$ \\ \hline
\citet{jader2018deep} & $\mathbf{94_{6}}$ & $84_{7}$ & $88_{5}$ \\ \hline
DeepOPG (Ours) & $90.7_{2.7}$ & $\mathbf{90.6_{2.0}}$ & $\mathbf{90.6_{1.8}}$ \\ \bottomrule
\end{tabular}
\begin{tabular}{ L{2.5cm}C{1.65cm}C{1.65cm}C{1.35cm}C{1.35cm}C{1.35cm}C{1.35cm} }
\toprule
\multirow{2}{*}[-0.5em]{\textbf{ Method}} & \multicolumn{4}{c}{\bf Natural Tooth Detection} & \multicolumn{2}{c}{\bf Implant Detection} \\ \cmidrule{2-5} \cmidrule(l){6-7}
  & Sensitivity\tablefootnote{\label{fn:tuzoff} They considered detection of teeth as 32 one-vs-all sub-problems. Even when a tooth is mis-labelled, it still is correct on 30 problems, and hence the high metrics.} (\%) & Precision\footnoteref{fn:tuzoff} (\%) & $\text{AP}_{0.5}$ (\%) & $\text{AP}_{0.7}$ (\%) & $\text{AP}_{0.5}$ (\%) & $\text{AP}_{0.7}$ (\%) \\ \midrule
\citet{tuzoff2019tooth} & $99.4$ & $99.4$ & $-$ & $-$ & $-$ & $-$ \\ \hline
\citet{kim2020automatic} & $-$ & $-$ & $96.7$ & $75.4$ & $45.1$ & $26.6$ \\ \hline
DeepOPG (Ours) & $\mathbf{100.0}$ & $\mathbf{99.8}$ & $\mathbf{97.6_{0.3}}$ & $\mathbf{89.4_{1.0}}$ & $\mathbf{75.0_{0.1}}$ & $\mathbf{75.0_{0.1}}$ \\ \bottomrule
\end{tabular}
\end{table}

\subsection{Comparing Existing Works}
\label{sec:comparing}
Comparison of model performances across works suffers from not only dataset difference but also clinical task difference. 
While we are unable to obtain proprietary datasets from previous works for evaluation, we can set up DeepOPG to similar settings to allow fairer comparisons.
For example, in Table~\ref{tbl:comparison}, \citep{wirtz2018automatic} and \citep{jader2018deep} tackled teeth-only segmentation, and hence we ignore error resulting from classes other than the teeth and the background in our segmentation module for a fair comparison.
\citep{tuzoff2019tooth} and \citep{kim2020automatic} addressed detection of natural teeth and implants, and thus we compare only detection results. 
Across all tasks except for precision in tooth segmentation, we are able to show superior performance.

Finally, for the \emph{missing teeth} finding summary, \citep{kim2020automatic} reached a sensitivity of 75.5\% and a precision of 84.5\% at a specificity of 80.4\%. 
Under the same specificity, we have a sensitivity of 94.3\% and a precision of 96.4\%.

%% file: 4-conclusion.tex
\section{Conclusion}
In this work, we provide an initial study, showing the possibilities to summarize findings for individual teeth from an orthopantomogram. 
By dividing the summarization process into two tasks: semantic segmentation and object detection, we can leverage weaker but faster-to-collect annotations to improve the detection model with reinforcement learning. 
The experiments demonstrate the efficacy of each module in the \this system, and, we hope to point the way for future works in this line and encourage dental imaging research.

%% file: appendix.tex
\appendix

\section{Training Details}
\label{sec:training-details}
We briefly describe the details of our implementation in this section.

All code implementations are in Tensorflow, run on four NVidia GTX 1080 Ti GPUs.
All model training incorporates augmentations including random brightness, contrast, affine transformation, elastic transformation, and Gaussian blurring.

\subsection{Functional Segmentation Module}
The U-Net model is trained with cross-entropy loss on the Adam~\citep{kingma2014adam} optimizer.
The learning rate is $10^{-5}$, the weight decay is $10^{-4}$, and the batch size is $4$. 
Models are train for 12,000 steps.

\subsection{Tooth Localization Module}
The Mask-RCNN~\citep{he2017mask} is trained similarly to the original work with the SGD optimizer.
The model is first trained with densely annotated masks only.
The learning rate is set to $10^{-3}$, the weight decay is $10^{-4}$, and the batch size is 1. 
Models are trained until 100,000 steps.

In the fine-tuning reinforcement learning step with DCR, we reduce the learning rate to $10^{-5}$ and only train the last network layer.
No weight decay is applied, and the models are trained for another 150,000 steps.
From each image, we draw 64 samples ($\hat\E$) from each set of detection output and baseline the rewards with the average reward across 64 samples.

\newpage

\section{Additional Results}
\label{sec:additional-results}

\begin{figure}[!h]
    \centering
    \includegraphics[width=0.6\textwidth]{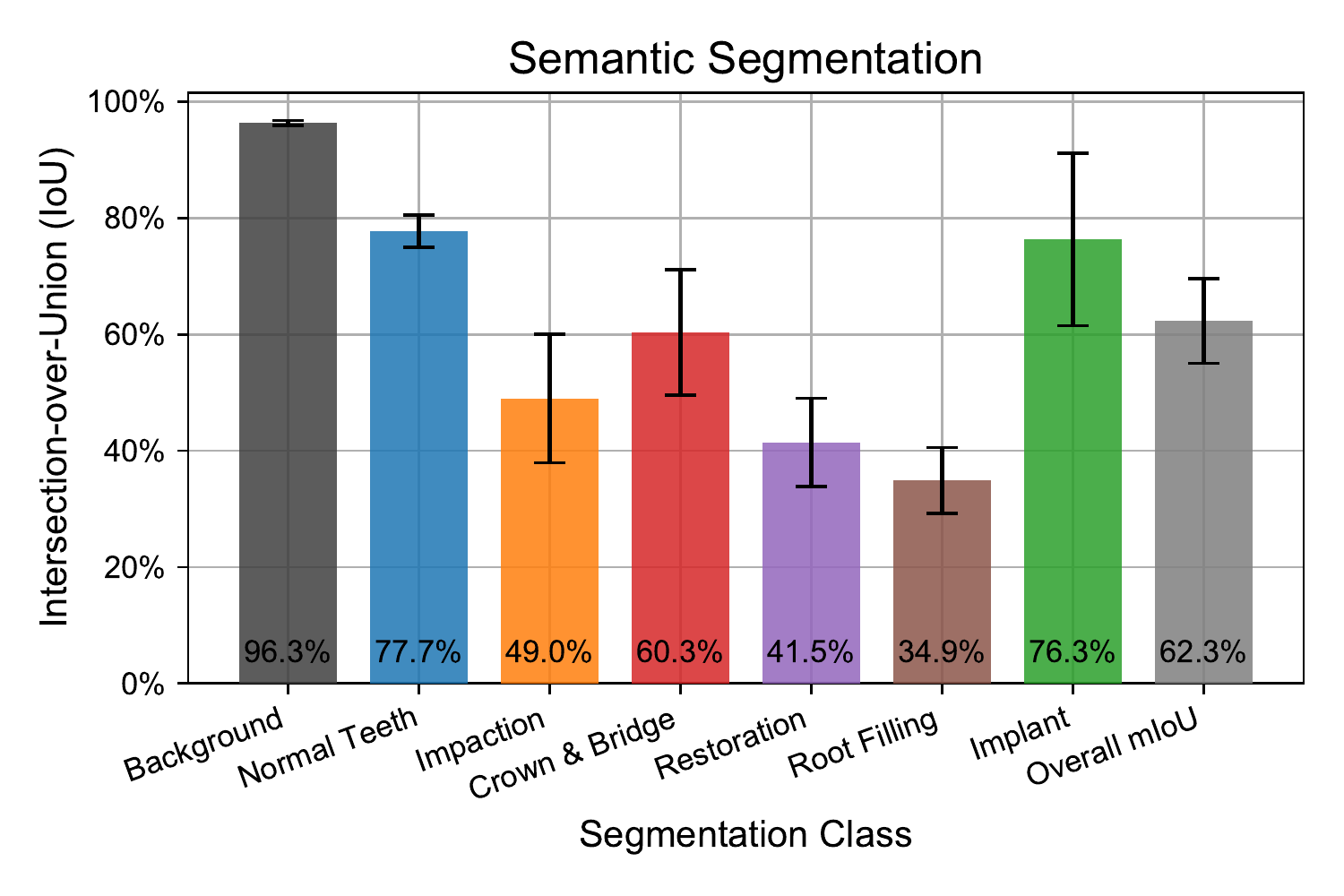}
    \caption{
        \textbf{Evaluation of the Functional Segmentation Module.}
        The IoU between the predicted segmentation and the ground truth segmentation are shown per class.
        Standard deviation is also labeled.
    }
    \label{fig:iou}
\end{figure}

\begin{figure}[!h]
    \centering
    \includegraphics[width=0.9\textwidth]{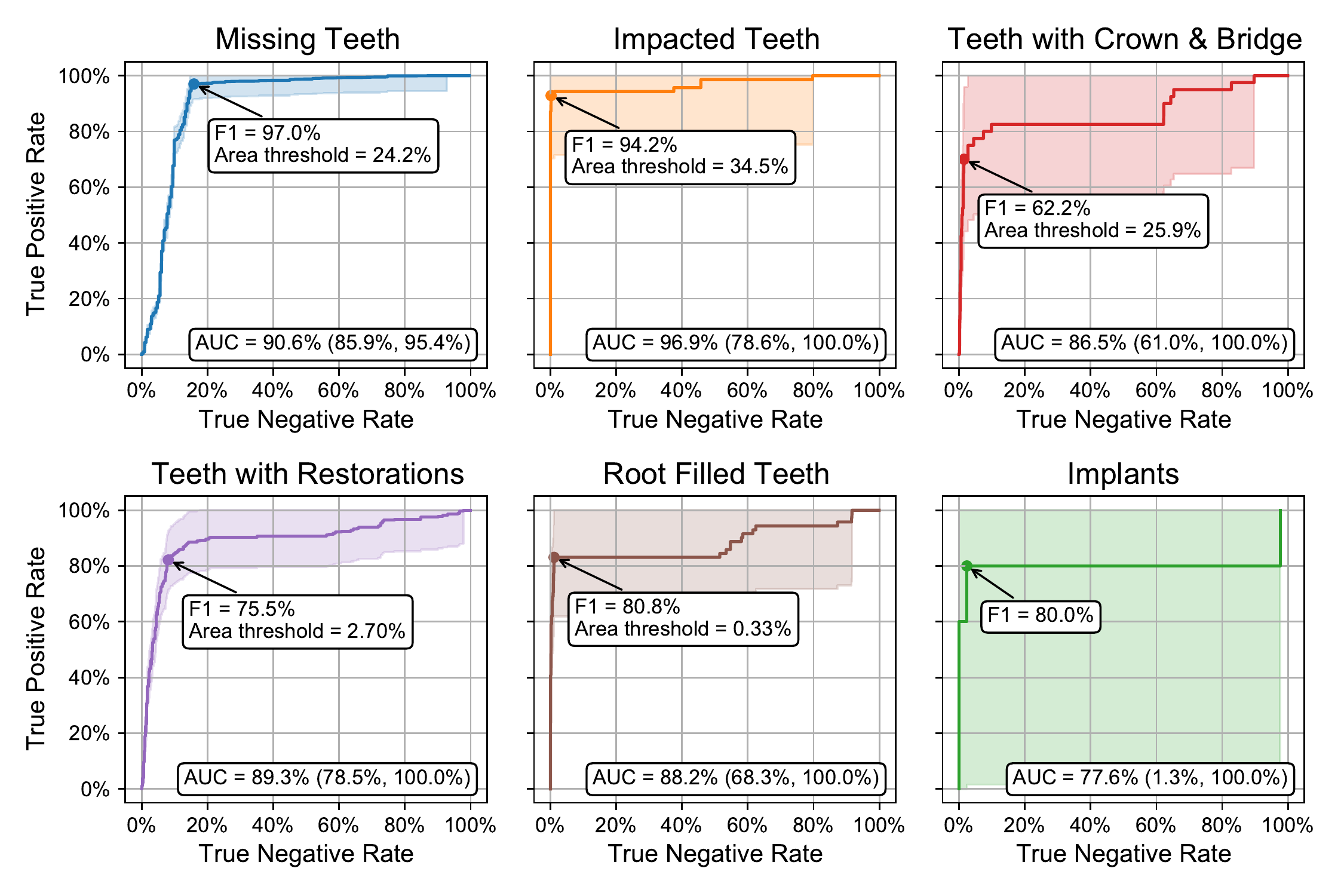}
    \caption{
        \textbf{Receiver Operating Characteristic (ROC) Curves.}
        Six finding types are considered.
        95\% confidence interval is annotated in shades and parentheses, and the operating points with the highest F1, along with the thresholds on percentage area, are labeled.
    }
    \label{fig:roc}
\end{figure}